\documentclass[letterpaper]{article} 
\usepackage{aaai23}  
\usepackage{times}  
\usepackage{helvet}  
\usepackage{courier}  
\usepackage[hyphens]{url}  
\usepackage{graphicx} 
\usepackage{natbib}  
\usepackage{caption} 
\usepackage{algorithm}
\usepackage{algorithmic}

\usepackage{newfloat}
\usepackage{listings}
\DeclareCaptionStyle{ruled}{labelfont=normalfont,labelsep=colon,strut=off} 
\floatstyle{ruled}
\newfloat{listing}{tb}{lst}{}
\floatname{listing}{Listing}

\usepackage{color}
\usepackage{multirow}
\usepackage{subcaption}
	
\usepackage{colortbl}
\usepackage{amsmath}
\usepackage{amssymb}
\usepackage{soul}

\definecolor{arylideyellow}{rgb}{0.91, 0.84, 0.42}
\definecolor{electriclavender}{rgb}{0.96, 0.73, 1.0}
\definecolor{forestgreen}{rgb}{0.13, 0.55, 0.13}

\newcommand{\decrease}[1]{\textcolor{red}{\hfill\scriptsize{$\downarrow$#1}}}
\newcommand{\decralgn}[1]{\textcolor{red}{\scriptsize{$\downarrow$#1}}}
\newcommand{\increase}[1]{\textcolor{forestgreen}{\hfill\footnotesize{$\uparrow$\textbf{#1}}}}

\definecolor{LightCyan}{rgb}{0.88,1,1}
\definecolor{Gray}{gray}{0.9}
\sethlcolor{LightCyan}
\definecolor{amber}{rgb}{1.0, 0.75, 0.0}
\definecolor{goldenpoppy}{rgb}{0.99, 0.76, 0.0}
\definecolor{inchworm}{rgb}{0.7, 0.93, 0.36}
 \definecolor{battleshipgrey}{rgb}{0.52, 0.52, 0.51}

\title{Contrastive Self-Supervised Learning\\Leads to Higher Adversarial Susceptibility}
\author{
    Rohit Gupta, \textsuperscript{\rm 1} 
    Naveed Akhtar, \textsuperscript{\rm 2} 
    Ajmal Mian, \textsuperscript{\rm 2} 
    Mubarak Shah \textsuperscript{\rm 1}
}
\affiliations{
    \textsuperscript{\rm 1}Center for Research in Computer Vision, University of Central Florida\\
    \textsuperscript{\rm 2} University of Western Australia\\

    rohitg@knights.ucf.edu, \{naveed.akhtar, ajmal.mian\}@uwa.edu.au,  shah@crcv.ucf.edu
}

\usepackage{bibentry}

\begin{document}

\maketitle

\begin{abstract}



Contrastive self-supervised learning (CSL) has managed to match or surpass the performance of supervised learning in image and video classification. However, it is still largely unknown if the nature of the representations induced by the two learning paradigms is similar. We investigate this under the lens of adversarial robustness. Our analysis of the problem reveals that CSL has intrinsically higher sensitivity to perturbations over supervised learning. We identify the uniform distribution of data representation over a unit hypersphere in the CSL representation space as the key contributor to this phenomenon. We establish that this is a result of the presence of false negative pairs in the training process, which increases model sensitivity to input perturbations. Our finding is supported by extensive experiments for image and video classification using adversarial perturbations and other input corruptions. We devise a strategy to detect and remove false negative pairs that is simple, yet effective in improving model robustness with CSL training. We close up to 68\% of the robustness gap between CSL and its supervised counterpart. Finally, we contribute to adversarial learning by incorporating our method in CSL. We demonstrate an average gain of about 5\% over two different state-of-the-art methods in this domain. 
\end{abstract}

\section{Introduction}

\begin{figure*}[t]
\begin{center}
  \includegraphics[width=0.9\textwidth]{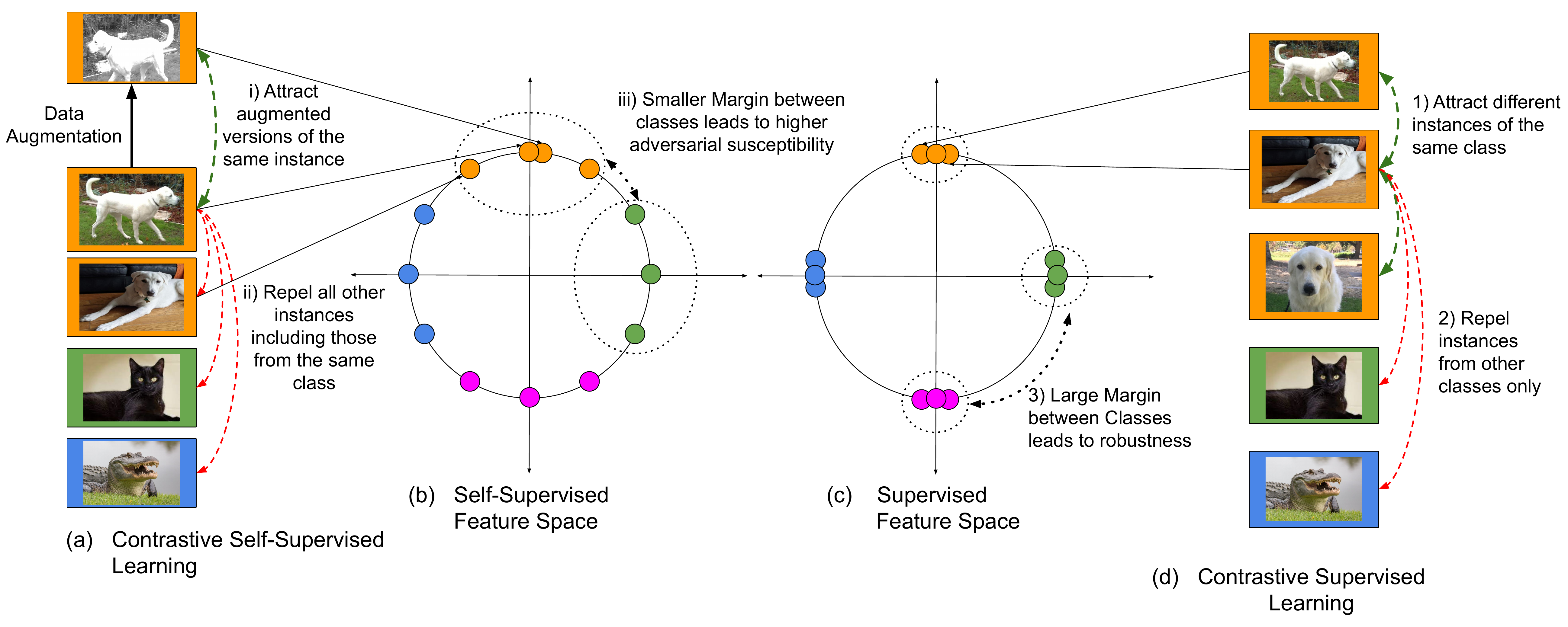}
\end{center}
  \caption{ \textbf{(a)} In Contrastive Self-supervised Learning (CSL) the anchor instance forms a positive pair with its augmented version, while being uniformly repelled from all other instances, including instances of the same class, which results in \textbf{(b)} a representation space with large class clusters and small inter-class margins. \textbf{(d)} In Supervised Contrastive Learning (SupCon) all instances of a given class are attracted to each other, while instances of different classes are repelled, which allows for each class to occupy a compact region of the feature space \textbf{(c)}, and as a result, has larger inter-class margins between class clusters. Lower inter-class margins in the self-supervised case lead to higher adversarial susceptibility. See Section~\ref{sec:explanation} for discussion.
  }
\label{fig:main}
\end{figure*}







Contrastive Self-supervised Learning (CSL)~\cite{chen2020simple} is a widely adopted technique of self-supervised training of visual models~\cite{clip}, \cite{mocov3}. It allows pre-training on unlabelled large-scale data prior to task-specific finetuning of the model. Since state-of-the-art CSL models in computer vision are performing equally well as the supervised models on popular benchmark datasets such as \texttt{ImageNet}, and make similar errors~\cite{geirhos2020surprising}, it is a common perception that CSL models learn similar representations as their  supervised counterparts.   


In this work, we scrutinize the under-investigated question of similarity between supervised and CSL representations from the perspective of adversarial robustness. Surprisingly, our findings do not align well with the prevailing belief that both model types admit representations of similar nature. We find that CSL models are considerably inferior to supervised models in terms of adversarial robustness in image and video classification tasks. We show that this disparity also holds when equivalent augmentations and training schedules are applied to both supervised and self-supervised learning.  

We investigate the reasons of higher adversarial vulnerability of CSL by examining the properties of the representation space induced by the contrastive loss. As illustrated in Fig.~\ref{fig:main}, Contrastive Learning (CL) brings similar pairs of data closer to each other in the learned representation space, and repels away the dissimilar pairs. In supervised CL, positive and negative pairs can be formed using known labels.  For the self-supervised case, typically an instance in the dataset is treated as a negative for every other instance, and positives for an instance are generated using data augmentation. 

\citet{wangisola} highlighted two key characteristics of  CSL representation. (a)~\textit{Alignment}: positive pairs formed by data augmentation of a single instance are closer in the feature space, and (b) \textit{uniformity}: a property that induces uniform distribution of instances in the representation space by repelling all other instances away from the anchor instance. The latter can be understood as an application of the principle of maximum entropy~\cite{jaynes1957information} (colloquially referred to as {\it Occam's Razor}). Since class information about the instances is not available in  self-supervised learning, it preserves  maximum information about the data in the representation by inducing a uniform distribution of the training  instances. 
Typically CSL representation lies on an n-dimensional hyper-sphere. As per the uniformity property,  instances in the training dataset are  roughly uniformly distributed over the surface of the hyper-sphere (Fig.~\ref{fig:main}b). However, this is sub-optimal from the perspective of adversarial robustness as this results in instances  being  close to the class boundaries and class clusters being spread out in the feature space. Contrastingly, in supervised learning, since the class labels are known, we can choose to only repel the instances from other classes to obtain tightly clustered classes in the feature space, achieving   higher adversarial robustness.


A solution to this problem is to identify instance pairs in the unlabelled training data that belong to the same class and avoid using them as negative pairs for the contrastive loss. However, doing so is not trivial because data labels are unavailable in self--supervised learning. We  propose an Adaptive False Negative Cancellation (Adaptive FNC) method that improves CSL training by gradually sifting out the likely false negative pairs.  We demonstrate  a significant adversarial robustness gain in the existing CSL methods RoCL and ACL with the proposed  technique.   Our key contributions are summarized below.

\begin{itemize}
\setlength\itemsep{0.02em}
    \item We provide the first systematic evidence of higher sensitivity of CSL to input perturbations with extensive experimental verification for image and video classification.
    \item We establish a clear connection between adversarial susceptibility of CSL models and the uniformity of their representation. We identify false negative pairs in model training as the key reason of higher model sensitivity.  
    \item Leveraging our insights, we devise a strategy to improve CSL robustness without adversarial training.
    \item We contribute to adversarial CSL by incorporating our findings into RoCL~\cite{kim2020adversarial} and ACL~\cite{jiang2020robust}, achieving consistent performance gain against adversarial attacks. 
\end{itemize}

\section{Related Work}
\label{sec:related}




\noindent \textbf{Contrastive learning:}
In self-supervised learning, Contrastive Learning (CL) is widely considered an effective paradigm. Within CL, SimCLR~\cite{chen2020simple} builds upon the prior work of MoCo (Momentum Contrastive learning)~\cite{moco}, Augmented Multiscale Deep InfoMax (AMDIM)~\cite{amdim}, and Contrastive Predictive Coding (CPC)~\cite{cpc} to develop its CL  pipeline. The pipeline includes data augmentations and a projection head to align the learned network representation during training. While the performance of SimCLR has been lately matched or exceeded by MoCov2~\cite{mocov2}, MoCov3~\cite{mocov3} and SimCLRv2~\cite{simclrv2}, the fundamental  structure of the CL framework remains similar in these works. Contrastive learning has also been successfully extended to action classification in videos~\cite{qian2020spatiotemporal},~\cite{dave2021tclr}, and image classification using Transformer architectures~\cite{mocov3}.

Many variants of CSL have been developed to improve  its performance on clean data. MoCHi~\cite{kalantidis2020hard} generates synthetic hard negatives in representation space, and NNCLR~\cite{dwibedi2021little} dynamically mines positive examples from different instances. Debiased CSL~\cite{chuang2020debiased} modifies the loss function to account for the presence of negative pairs formed by different instances belonging to the same class. FNC~\cite{huynh2022boosting} and IFND~\cite{chen2021incremental}  attempt to directly  identify and remove negative pairs during training. This concept is related to our approach, however, we focus on improving adversarial robustness whereas these prior works focus on improving accuracy on clean data.

Even though SwAV~\cite{caron2020unsupervised} does not use contrastive loss, it preserve the `alignment' property of its representation by clustering the augmented versions of instances. Moreover, it is also able to preserve the `uniformity' property by enforcing an explicit equi-partitioning constraint over its representation space. Recent non-CSL methods include SimSiam~\cite{SimSiam}, BYOL~\cite{BYOL} and DINO~\cite{dino}, which are self-distillation based methods.  These techniques do not use negative contrastive pairs and hence are not directly affected by the weaknesses of CSL. 


Owing to the promising performance of CL, recent works have also focused on exploring the unique properties of contrastive learning. Geirhos et al. \cite{geirhos2020surprising} found that such models produce results similar to those learned with supervision. \citet{xiao2021what} showed that the choice of the best data augmentation method for self-supervised training depends on the specific dataset. \citet{purushwalkam2020} claimed that CL  results in superior occlusion-invariant representations, while \citet{wangisola} analyzed  CL by studying the alignment and uniformity properties of feature distribution. These properties are claimed to endow more discriminative power to the  models. The uniformity property of CL is also discussed by \citet{chen2020intriguing}, who refer to it as the `distribution' property. \citet{Wang_2021_CVPR}  built a relationship between the uniformity and the temperature hyper-parameter of the loss function.

\noindent{\bf Robustness and self-supervision:} In prior art on  robustification of supervised learning,   self-supervision is considered as a helpful tool. \citet{hendrycks2019} found that adversarial robustness of supervised models can be improved by adding an additional self-supervised task in a multi-task approach. Similarly, \citet{carmon2019unlabeled} also found that using additional unlabeled data improves adversarial robustness of the model. \citet{chen2020adversarial} also developed robust versions of pretext-based self-supervised learning tasks and demonstrated that this, along with robust fine-tuning of the model, results in significant improvement in the robustness relative to the baseline adversarial training. 

\noindent{\bf Adversarial training for self-supervised models:} 
Efforts have also been made for adversarial training  of self-supervised learning. \citet{kim2020adversarial} developed an instance-based adversarial attack for contrastive self-supervised training,  and used it for adversarial training. The concurrent work by \citet{Jiang2020} develops an adversarial CL  framework that is claimed to surpass prior self-supervised learning methods in robustness as well as accuracy on clean data. \citet{Ho2020} created a generalized formulation of AdvProp training~\cite{Xie_2020_CVPR} applicable to self-supervised learning, with the goal to increase accuracy on clean data. Adversarial training  increases robustness of CL models, but incurs a significant training cost. Hence, we contribute to  CSL robustness without resorting to adversarial training. We also show that adversarial training can further benefit from our technique. 

\begin{figure*}[t]
\begin{center}
\begin{subfigure}[b]{0.48\linewidth}
         \centering
         \begin{minipage}{0.32\linewidth}
             \includegraphics[width=\linewidth]{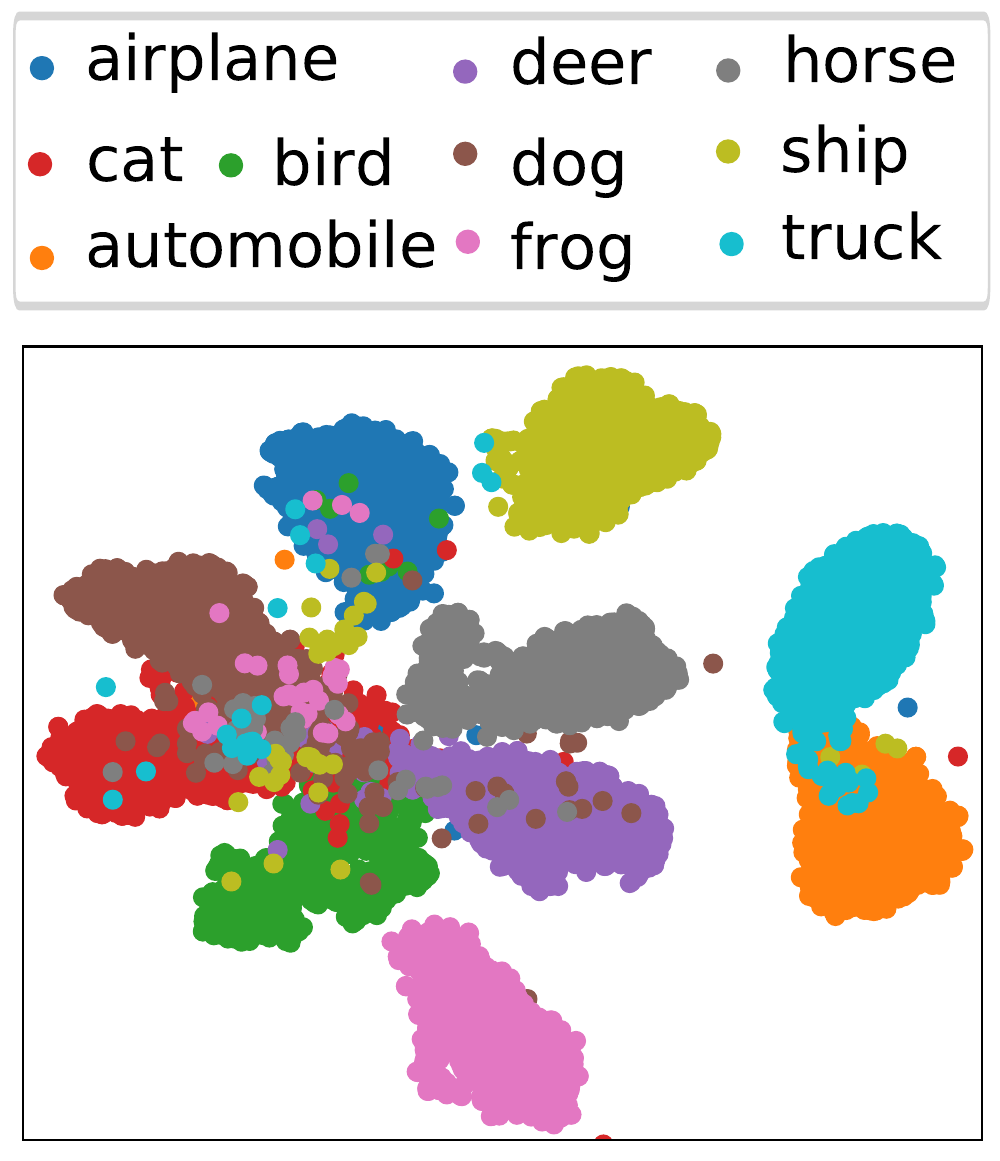}
         \end{minipage}
         \begin{minipage}{0.65\linewidth}
             \includegraphics[width=\linewidth, trim=4mm 4mm 6mm 10mm, clip]{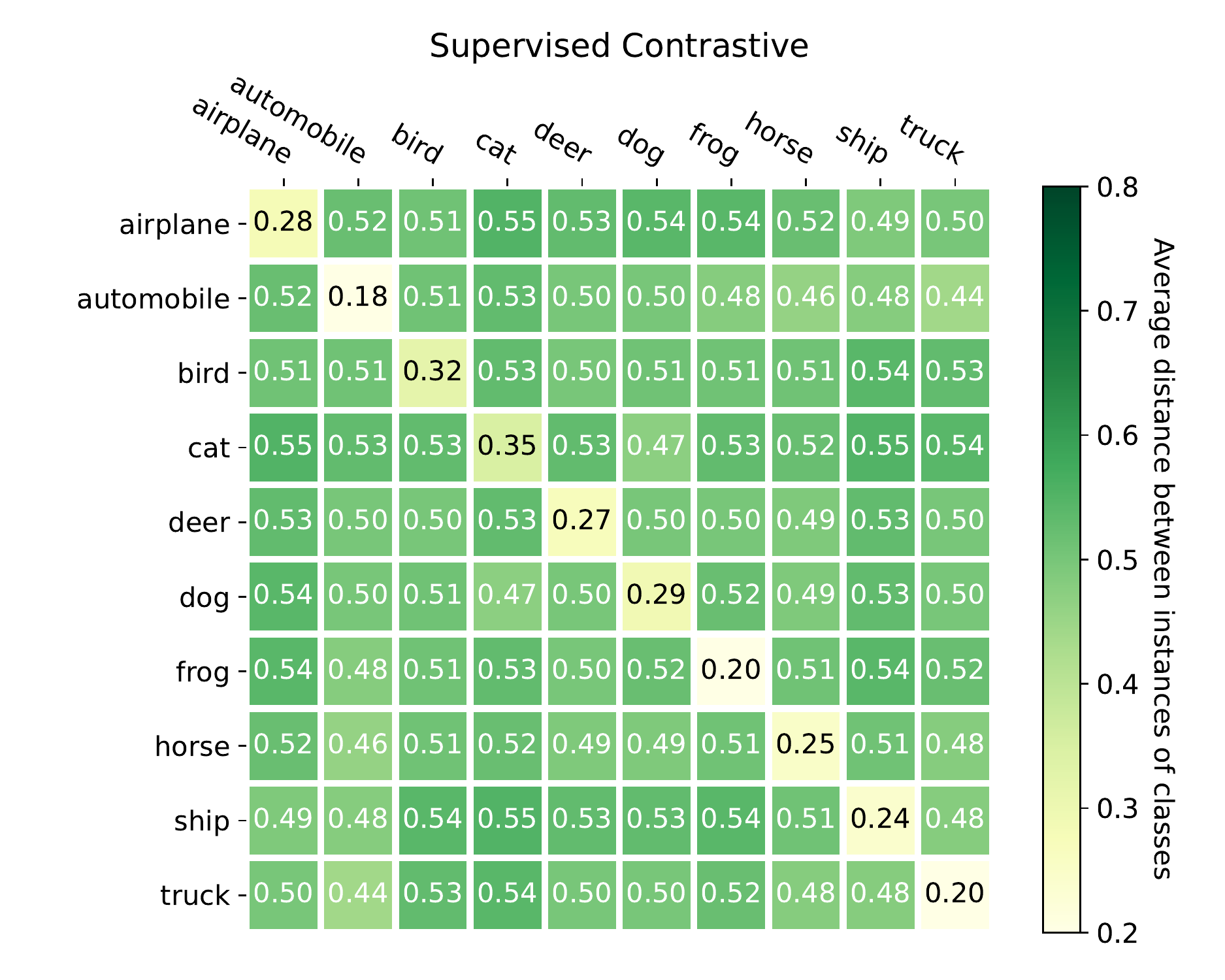}
         \end{minipage}
         \caption{Supervised}
         \label{fig:supactual}
     \end{subfigure}
     \quad
    \begin{subfigure}[b]{0.48\linewidth}
        \centering
        \begin{minipage}{0.32\linewidth}
           \includegraphics[width=\linewidth]{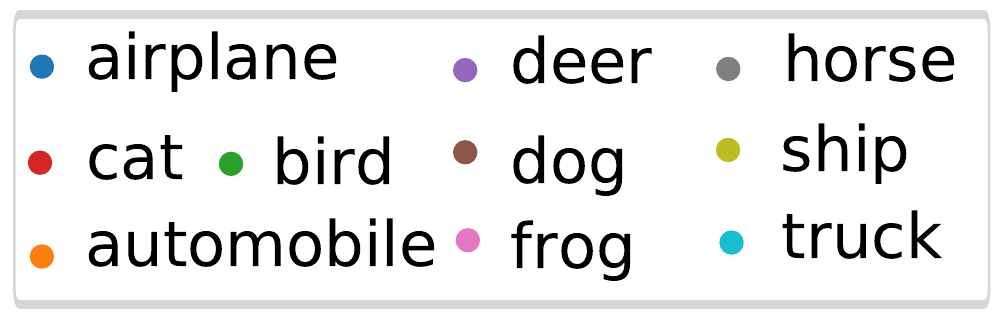}\\
           \includegraphics[width=\linewidth, trim=0mm 0mm 0mm 0mm, clip]{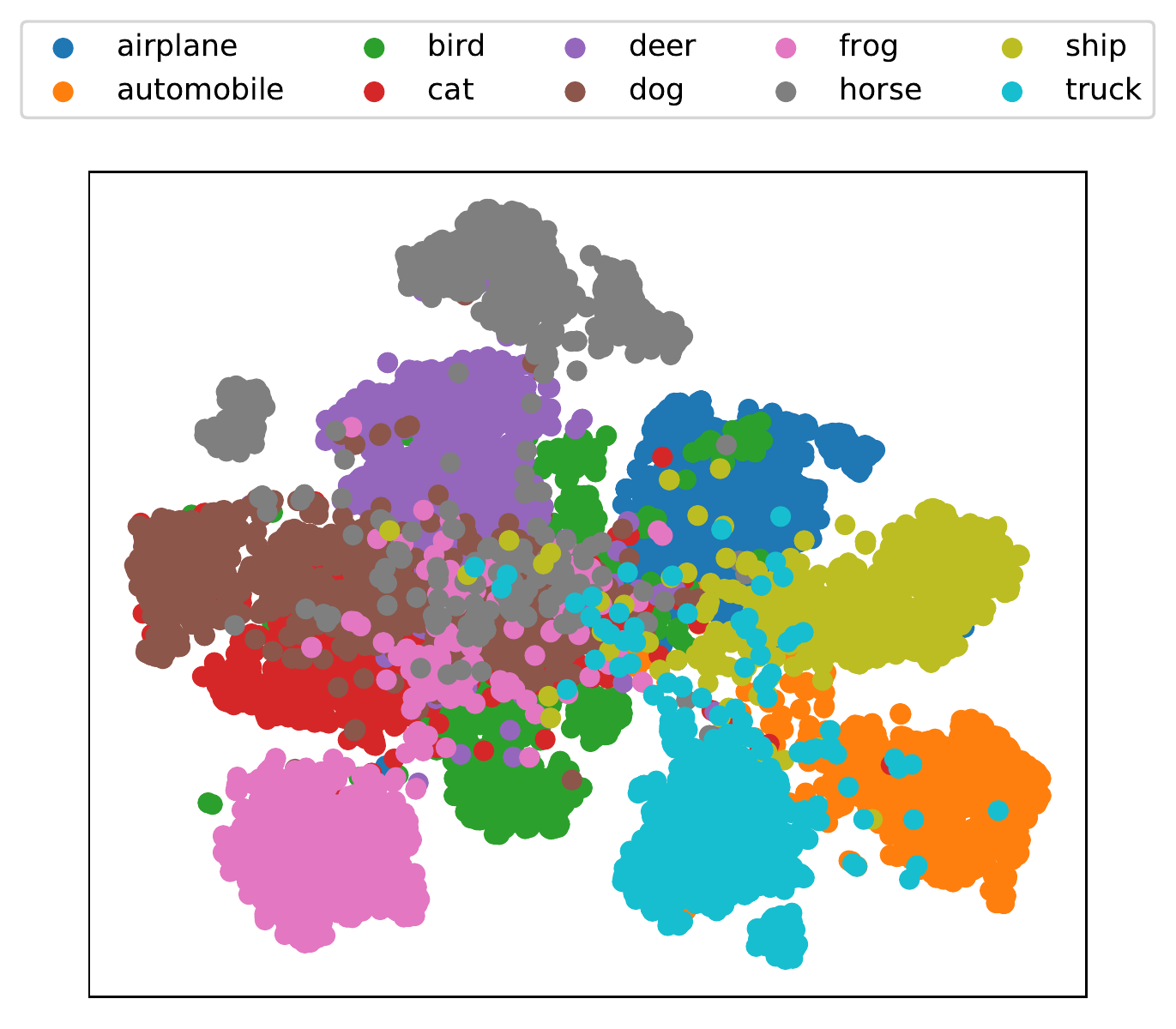}
        \end{minipage}
        \begin{minipage}{0.65\linewidth}
            \includegraphics[width=\linewidth, trim=4mm 4mm 6mm 10mm, clip]{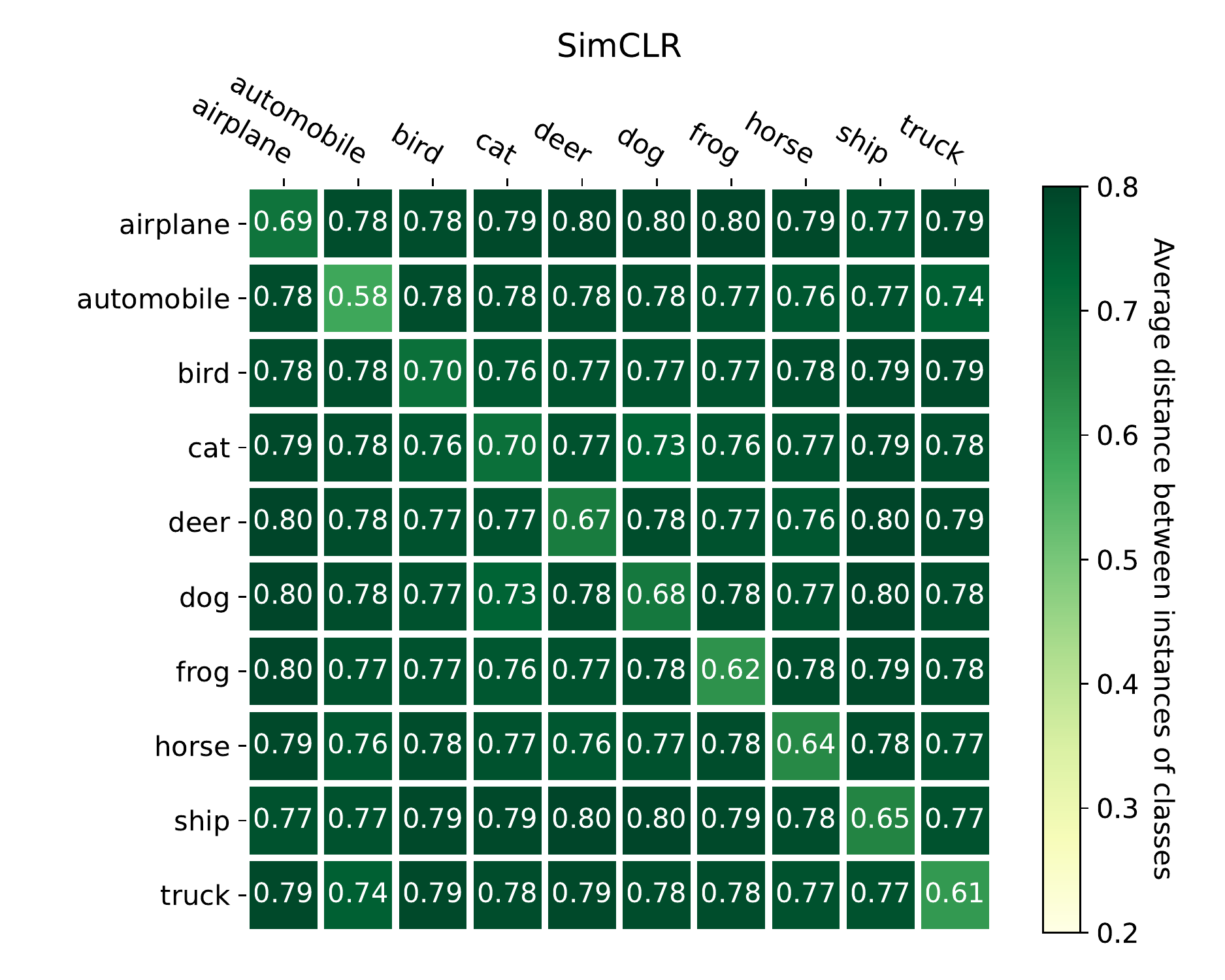}
        \end{minipage}
         \caption{Self-Supervised}
         \label{fig:selfsupactual}
     \end{subfigure}
     

\end{center}
  \caption{tSNE Visualization of representation space and average inter- and intra- class distances for CIFAR-10 instance pairs obtained with \textbf{(a)}~Supervised, and \textbf{(b)}~Self-Supervised contrastive learning. Average ratio of inter-class distances relative to intra-class distances is much lower for the Self-Supervised model  (1.19$\times$) than for Supervised (1.98$\times$), which leads to higher adversarial susceptibility for the self-supervised model.}

\label{fig:actualfeatures}
\end{figure*}

\begin{table*}[h]
\centering
{\small
\bgroup
\def\arraystretch{1.1}
\resizebox{\textwidth}{!}{%
\begin{tabular}{lccc|cc|c|c}
\hline

\hline

\hline
\textbf{Pre-Training} & & \multicolumn{2}{c|}{\textbf{FGSM-$\ell_{\infty}$}}  & \multicolumn{2}{c|}{\textbf{PGD-$\ell_2$}} & \textbf{PGD-$\ell_1$} & \texttt{\textbf{AutoAttack}}-$\ell_\infty$ \\ 
\cline{2-8}
& $\epsilon$ = \textbf{0} & $\epsilon$ = \textbf{8/255} & $\epsilon$ = \textbf{16/255} & $\epsilon$ = \textbf{0.1} & $\epsilon$ = \textbf{0.25} & $\epsilon$ = \textbf{8.0} & $\epsilon$ = \textbf{1/255}  \\
\hline
\multicolumn{8}{c}{\textbf{CIFAR-10}} \\
\hline 
Supervised (Cross entropy)    & 95.4   & 29.6 \decrease{69\%} & 20.1 \decrease{79\%}& 31.2\decrease{67\%} & 15.7\decrease{84\%} & 16.5\decrease{83\%} & 18.6 \decralgn{81\%}  \\
Supervised (Contrastive)      & 95.5   & 38.8 \decrease{59\%} & 31.8 \decrease{67\%}& 34.2\decrease{64\%} & 18.4\decrease{81\%} & 20.7\decrease{78\%} & 24.3 \decralgn{75\%}  \\
\rowcolor{LightCyan}
Self-Supervised (Contrastive) & 92.7   & 26.8 \decrease{71\%} & 13.4 \decrease{86\%}& 20.9\decrease{77\%} &  8.3\decrease{91\%} & 11.5\decrease{88\%} & 13.7 \decralgn{85\%}  \\
\hline
 \multicolumn{8}{c}{\textbf{CIFAR-100}} \\
\hline 
Supervised (Cross entropy)    & 74.9  & 14.3 \decrease{81\%} & 8.4 \decrease{89\%} & 23.1 \decrease{69\%} & 11.5\decrease{85\%} & 12.1\decrease{84\%} & 13.6 \decralgn{82\%} \\
Supervised (Contrastive)      & 76.3  & 12.6 \decrease{83\%} & 6.7 \decrease{91\%} & 21.9 \decrease{71\%} & 9.2 \decrease{88\%} & 13.4\decrease{82\%} & 11.9 \decralgn{84\%} \\
\rowcolor{LightCyan}
Self-Supervised (Contrastive) & 68.9  & 9.40 \decrease{87\%} & 3.0 \decrease{96\%} & 13.7 \decrease{80\%} & 4.4 \decrease{94\%} &  6.8\decrease{90\%} & 8.7 \decralgn{88\%} \\
\hline
\end{tabular}}
\egroup
}
\caption{Supervised and CSL CIFAR models \hl{(highlighted)} are trained with similar training setups and their robustness is compared for FGSM, \texttt{AutoAttack} (both $\ell_{\infty}$) and PGD variants ($\ell_1$ \& $\ell_2$).}
\label{tab:CIFARFGSM}
\end{table*}

\section{Adversarial Susceptibility of CSL}
\label{sec:explanation}


The popular contrastive self-supervised  representation learning strategy, e.g.,~used by SimCLR~\cite{chen2020simple}, learns a representation space from unlabeled data. It samples the so-called  `positive pairs' by applying  independent random transformations to an original sample (a.k.a.~anchor). The positive pairs are expected to have representations similar to the anchor. The `negative pairs' are formed by pairing the anchor with other original instances. If $x$ represents the anchor, $x^+$ represents the positive pairing generated by augmentation and $x^-_i$s represent the other instances in the training batch, then the contrastive loss,  $ \mathcal L_{c}(x)$, is given as
\begin{align}
    \mathcal L_{c}(x) \!=\!\! {\underset{x, x^+, \{x^-_i\}^N_i}{\mathbb E}} \! \Big[-\log\frac{sim(x,x^+)}{sim(x,x^+) + \sum_i^N sim(x,x_i^-)}\Big],
    \label{eq:loss}
\end{align}
 For brevity in Eq.(~\ref{eq:loss}) we define $sim(x,y)$ as $e^{f(x)^T \cdot f(y)/\tau}$, which is the normalized temperature scaled similarity between the representations of inputs $x$ and $y$.  Here, $f(x)$ is the representation of a given input, which is constrained by design to lie on a $d$ dimensional hypersphere. 

In order to link CSL to adversarial susceptibility, we make following three claims:

\begin{enumerate}
\setlength\itemsep{0.02em}
    \item Presence of false negative pairs in CSL leads to \emph{instance-level} uniformity
    \item Instance level uniformity implies weaker separation of classes in the feature space (lower ratio of average inter-class to intra-class distance) 
    \item Weaker separation of classes results in higher adversarial susceptibility 
\end{enumerate}

Firstly, since all instances in CSL loss repel each other, it is  likely (illustrated in Figure~\ref{fig:main}) that instance level uniformity is the stable equilibrium to minimize the loss. However, prior work~\cite{wangisola} also provides an analytical proof that  in the limit of large batch size ($N \rightarrow \infty$) the contrastive loss is equivalent to reducing:

\begin{align}
    \mathcal{L}_c'(x) \equiv -\frac{1}{\tau}{\underset{x, x^+}{\mathbb{E}}} sim(x,x^+) + {\underset{x, x^-}{\mathbb{E}}}\log sim(x^-,x).
    \label{eq:uniformity}
\end{align}

Here, the first term ensures \textit{alignment} of positive pairs, whereas the second term ensures \textit{uniformity} of instances over the hypersphere. Note that in the case of supervised learning, since instances of the same class are not repelled, there is no instance level uniformity. Rather, the uniformity term applies at the \textit{class-level}.

Which brings us to our second claim. We know from geometry that the surface area of a d-dimensional sphere is given by: $\text{A}(d) = \left(2 \pi^{\frac{d-1}{2}} / \Gamma (\frac{d-1}{2}  )\right)$, where $\Gamma(x)$ is the gamma function. This surface area is divided very differently in supervised and self-supervised learning. Since the classes as uniformly spread out, the distance between a class center and the nearest boundary should be proportional to $A_c = \frac{\text{A}(d)}{C}$, where $C$ is the number of classes in the dataset. In the supervised case, as training progresses, instances of a class will come closer and closer to each other. We choose to indicate this minimal intra-class separation by $s_{min}$.  Whereas in the self-supervised case, the intra-class margin has to be higher than $\frac{\text{A}(d)}{N}$ due to instance level uniformity. If we define a measure of class separation in terms of the ratio of the average inter-class distance to the average intra-class distance, we would expect $\rho_{SupCon} = \frac{A_c}{s_{min}}$  and $\rho_{CSL} = \frac{A_c \times N}{C}$. Since $s_{min} \rightarrow 0$ and $N/C \in \text{finite rationals } \mathbb{Q}$, which implies $\rho_{SupCon} > \rho_{CSL}$. In Section~\ref{sec:experiments} we also provide empirical evidence to show that intra-class distances are much smaller for the supervised models than for CSL.

Finally, we come to our third claim, which asserts that classifiers with stronger separation of classes in feature space are more robust. This follows intuitively from the definition of adversarial examples. Adversarial examples exist because clean examples from the dataset lie very close to decision boundaries in feature space and can be pushed across them with a small perturbation. Prior works such as DeepFool~\cite{moosavi2016deepfool} define classifier robustness in terms of the average minimum perturbation magnitude required to shift the labels for the dataset. Since deep classifiers are typically smooth, a larger magnitude of perturbation required in pixel space indicates higher distance of instance from nearest decision boundary.





\section{Empirical Verification of Susceptibility}
\label{sec:experiments}


In order to empirically verify our finding of higher susceptibility for CSL, we compare robustness of supervised and self-supervised image and video classification models. First, we train models with identical architectures and comparable training policies on the CIFAR-10 and CIFAR-100 datasets in order to isolate the effect of CSL training on robustness. Second, we demonstrate that our finding of adversarial susceptibility of CSL also extends to ImageNet pre-trained models.  Third, we also demonstrate that to analyze the robustness of models to image transformations e.g blurring and noise addition. Finally, we move to the task of video classification and demonstrate that our observations also apply to models trained on UCF101 and HMDB51 datasets, two widely used benchmarks for action recognition.


 For comparing the robustness of models (which can have varying clean accuracy) we use the drop in accuracy relative to the clean accuracy:  

 \begin{equation} 
\mathcal{P}_{Drop} = \frac{\mathcal{P}(y|x) - \mathcal{P}(y|x+\Delta x)}{\mathcal{P}(y|x)},
\end{equation}

\noindent where $\mathcal{P}(y|x)$ is the model accuracy for clean data and   $\mathcal{P}(y|x+\Delta x)$ is the accuracy with perturbed data.  $x$ and $y$ represent data and correct label respectively.

\begin{table*}[t]
\centering
{\small
\begin{tabular}{ll|c|cc|cc|c|c}
\hline

\hline

\hline
\multirow{2}{*}{\textbf{Model}} & \multirow{2}{*}{\textbf{Method}} & & \multicolumn{2}{|c|}{\textbf{FGSM}} & \multicolumn{2}{|c|}{\textbf{PGD}-$\ell_{\infty}$} & \textbf{PGD}-$\ell_{2}$ & \texttt{\textbf{AutoAttack}}-$\ell_\infty$ \\ 
\cline{3-9}
& & $\epsilon$ = \textbf{0} 
& $\epsilon$ = \textbf{.25/255} & $\epsilon$ = \textbf{1/255} 
& \textbf{$\epsilon$ = .25/255} & \textbf{$\epsilon$ = 1/255}  
& $\mathbf{\epsilon}$\textbf{ = 0.5} & \textbf{$\epsilon$ = .25/255}   \\
\hline
 \cellcolor{white} & Supervised  & 76.71 & 38.20 \decrease{50\%} & 11.53\decrease{85\%} & 28.22 \decrease{63\%}  & 0.65 \decrease{99\%}  & 11.3 \decrease{85\%}  & 16.16\decralgn{79\%} \\
\rowcolor{LightCyan}
 \cellcolor{white} & SimCLR      & 68.95   & 24.33 \decrease{65\%} & 8.85 \decrease{87\%} & 10.89 \decrease{84\%}  & 0.24 \decrease{100\%} & 5.2\decrease{92\%} &  4.41\decralgn{94\%} \\
\rowcolor{LightCyan}
 \cellcolor{white} & SwAV        & 75.34   & 23.35 \decrease{69\%} & 5.95 \decrease{92\%} & 11.73 \decrease{84\%}  & 0.20 \decrease{100\%} & 4.1  \decrease{95\%} &  10.81\decralgn{86\%} \\
 \rowcolor{inchworm}  \cellcolor{white} \multirow{-4 }{*}{ResNet50} & BYOL        & 74.56 & 39.40 \decrease{47\%}  & 13.50\decrease{82\%} & 32.61\decrease{56\%} & 1.89\decrease{97\%}  &   12.51\decrease{83\%} & 17.34\decralgn{77\%}  \\
\hline
 
 \cellcolor{white} & Supervised    & 81.07 & 61.44\decrease{24\%}  & 43.38\decrease{46\%} & 59.57\decrease{27\%} & 27.53\decrease{66\%}  & 42.2\decrease{48\%} & 46.99~\decralgn{42\%} \\
\rowcolor{LightCyan} 
 \cellcolor{white} & MoCoV3        & 76.66 & 46.69\decrease{39\%}  & 13.21\decrease{83\%} & 39.89\decrease{48\%} & 2.42\decrease{97\%}   & 19.6\decrease{75\%} & 27.88~\decralgn{64\%}  \\
\rowcolor{inchworm}  \cellcolor{white} \multirow{-3}{*}{ViT-B/16}  & DINO & 77.99 & 58.36\decrease{25\%}  & 22.17\decrease{72\%} & 57.95\decrease{26\%} & 14.14\decrease{82\%}  & 36.6\decrease{53\%} & 51.46~\decralgn{34 \%}   \\
\hline
\end{tabular}
}
\caption{ImageNet top-1 accuracy for various pre-trained models under FGSM, PGD and \texttt{AutoAttack} attacks. Percentage drop relative to clean input accuracy is given in \decralgn{red}. Contrastive Self-Supervised models \hl{(highlighted)} show higher drops.  
}
\label{tab:ImageNetFGSM}
\end{table*}

\subsection{Susceptibility of CIFAR Models}

We perform controlled experiments with supervised cross-entropy, supervised contrastive learning~\cite{Khosla2020SCL} and CSL using CIFAR-10 and CIFAR-100 datasets. While our argument on higher sensitivity of contrastive self-supervised learning is best verified using a weaker attack like FGSM, we also investigate the adversarial susceptibility of models to stronger attacks, e.g., Projected Gradient Descent (PGD)~\cite{madry2018towards} and AutoAttack~\cite{croce2020reliable}. AutoAttack is a composite of 4 attacks: untargeted and targeted step-size free Automatic PGD, Fast Adaptive Boundary Attack~\cite{croce2020minimally} and the black-box Square Attack~\cite{ACFH2020Square}. In order to ensure a fair comparison, all our ResNet50 models are trained for 1,000 epochs each, which ensures full convergence. We also use the same data augmentation strategy for all models, with the minor difference of using weaker color jittering for the supervised cross-entropy loss, as suggested by~\cite{chen2020simple}. To ensure reliability of results, they are averaged over 5 training runs.

The results of our experiments are summarized in Table~\ref{tab:CIFARFGSM}, which suggest that supervised models with contrastive loss are more resilient to adversarial manipulation as compared to their self-supervised counterparts. In Fig.~\ref{fig:actualfeatures}, we provide tSNE visualisation of the representations for supervised and self-supervised CIFAR-10 models learned under the contrastive loss in Table~\ref{tab:CIFARFGSM}. As can be observed from the respective class heatmaps, the ratio of inter-class to intra-class margin is much higher for supervised (1.98$\times$) than in the self-supervised case (1.19$\times$). Clearly, the supervised model is able to separate the classes better than the self-supervised model. This observation is in line with our finding that uniform representations in self-supervised contrastive learning renders the model more sensitive to input perturbations.

\subsection{Susceptibility of ImageNet Models}

 In order verify our results for widely deployed pre-trained ImageNet models, we focus on two popular architectures: ResNet50 and Vision Transformer (ViT-B). We compare the robustness of supervised models against models trained with self-supervision techniques. We select both CSL and non-CSL self-supervision methods in order to demonstrate the validity of our argument. SimCLR~\cite{chen2020simple}, a simple CSL method, and MoCov3, a state of the art technique which utilizes an additional momentum encoder and achieves better results are our primary focus. We also include SwAV~\cite{caron2020unsupervised}, which does not use contrastive loss, however, it also preserves the uniformity property of representations, which, according to our analysis in Section~\ref{sec:explanation}, is the primary cause of higher adversarial susceptibility. Finally, we test BYOL and DINO, which are a self-supervised method based on self-distillation and do not utilize negative pairs.  We maintain architectural similarity between different techniques to ensure comparable results. Along with adversarial attacks, we also study the robustness of models w.r.t.~other transformations e.g.,~adding noise, blurring, simulated fog etc.,~using ImageNet-C dataset.

 \begin{figure}[h]
\centering
  \includegraphics[width=\linewidth, trim={0 1.25cm 0 0.5cm},clip]{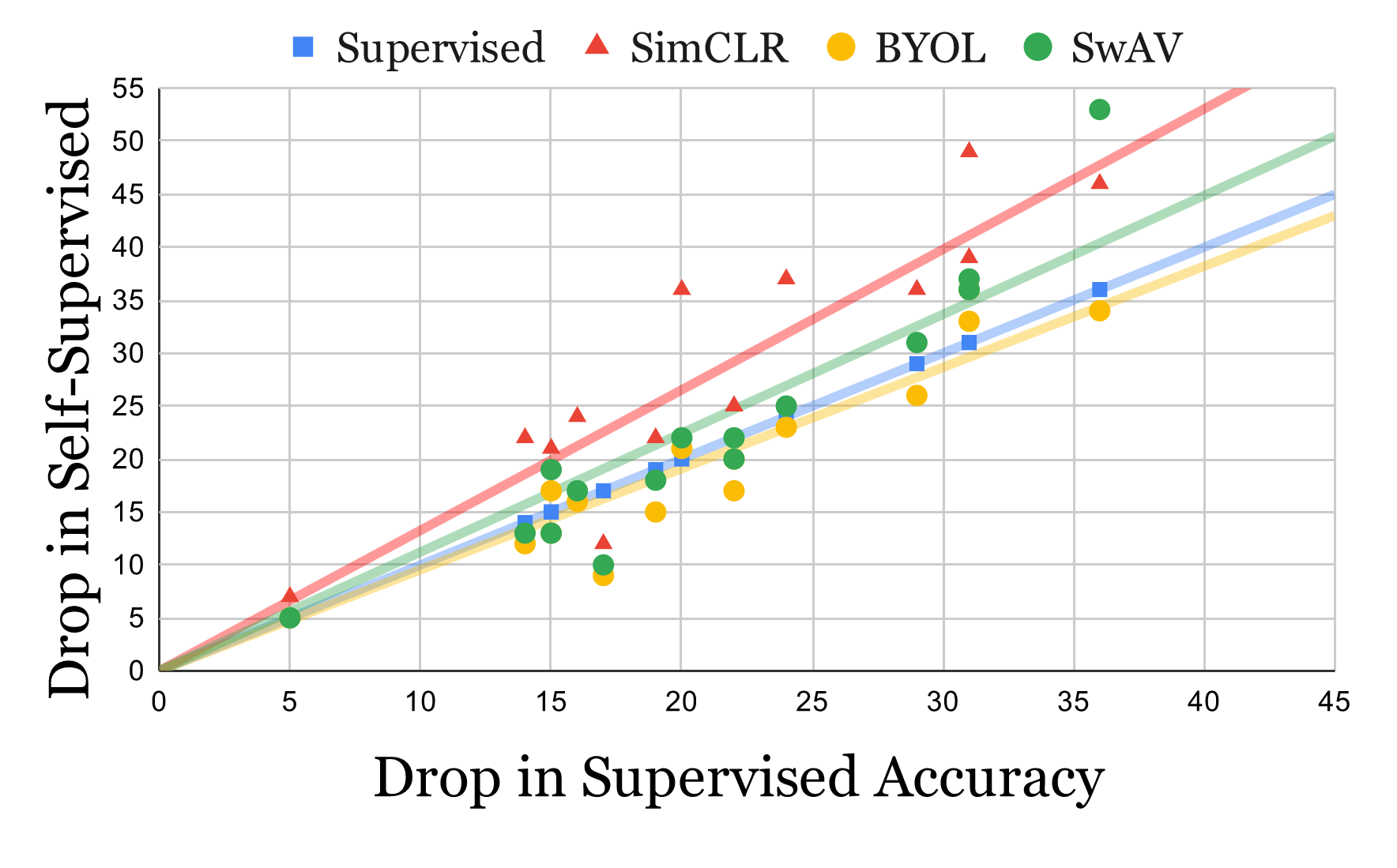}
  \caption{Relative accuracy drop (ResNet50) due to ImageNet-C corruptions. CSL (\textcolor{red}{\hl{\textbf{SimCLR}}}) has bigger drops than \textcolor{blue}{\hl{\textbf{Supervised}}} or SSL without negative pairs (\textcolor{amber}{\hl{\textbf{BYOL}}}).} 
\label{fig:drops}
\end{figure}

\begin{table*}
\centering
{\small
\bgroup
\def\arraystretch{1.0}
\begin{tabular}{l|c|ccc|cc|c}
\hline

\hline

\hline
& & \multicolumn{3}{c}{\textbf{FGSM-$\ell_{\infty}$}} & \multicolumn{2}{c}{\textbf{PGD-$\ell_{\infty}$}} & \texttt{\textbf{AutoAttack-$\ell_{\infty}$}} \\
\cline{3-8}
\textbf{Pre-Training}  & $\epsilon$ = \textbf{0} & $\epsilon$ = \textbf{1/255} & $\epsilon$ = \textbf{2/255} & $\epsilon$ = \textbf{4/255} & $\epsilon$ = \textbf{1/255} & $\epsilon$ = \textbf{2/255} & $\epsilon$ = \textbf{1/255} \\
\hline
 \multicolumn{8}{c}{\textbf{UCF101}} \\
 \hline
Supervised                         & 59.4  & 26.6 \decrease{55\%}  & 12.2 \decrease{79\%} & 3.9 \decrease{93\%}   & 15.6 \decrease{74\%}  & 5.2 \decrease{91\%} & 12.6 \decralgn{79\%} \\
\rowcolor{LightCyan}
TCLR~\cite{dave2021tclr}           & 75.5  & 10.8 \decrease{86\%}  &  6.1 \decrease{92\%} & 3.3 \decrease{96\%}   & 6.30 \decrease{92\%}  & 3.1 \decrease{96\%} & 4.00 \decralgn{95\%} \\
\rowcolor{LightCyan}
CVRL~\cite{qian2020spatiotemporal} & 60.2  &  6.00 \decrease{90\%}  &  3.0 \decrease{95\%} & 1.6 \decrease{97\%}  & 4.70 \decrease{92\%}  & 1.9 \decrease{97\%} & 3.70 \decralgn{94\%} \\
\hline
 \multicolumn{8}{c}{\textbf{HMDB51}} \\
 \hline
Supervised                         & 24.8  & 12.2 \decrease{51\%}  & 6.4 \decrease{74\%} & 3.0 \decrease{88\%}   & 7.2 \decrease{71\%}  & 2.7 \decrease{89\%} & 6.2 \decralgn{75\%} \\
\rowcolor{LightCyan}
TCLR~\cite{dave2021tclr}           & 47.6  & 5.7 \decrease{77\%}  &  2.7\decrease{89\%} & 1.7 \decrease{93\%}   & 2.7 \decrease{89\%}  & 2.0 \decrease{92\%} & 2.0 \decralgn{92\%} \\
\rowcolor{LightCyan}
CVRL~\cite{qian2020spatiotemporal} & 35.2  &  3.7 \decrease{85\%}  &  1.7 \decrease{93\%} & 0.5 \decrease{98\%}  & 2.2 \decrease{91\%}  & 1.0 \decrease{96\%} & 1.7 \decralgn{93\%} \\
\hline
\end{tabular}
\egroup
}
\caption{Top-1 accuracy for pre-trained UCF101 and HMDB51 video classification under FGSM, PGD and \texttt{AutoAttack} attacks. Percentage drop relative to clean data accuracy is given in \textcolor{red}{$\downarrow$red}. CSL methods TCLR and CVRL \hl{(highlighted)} have lower robust accuracy than the supervised model despite heavily outperforming supervised learning in clean accuracy.} 
\label{tab:UCFFGSM}
\end{table*}

 \noindent{\textbf{Susceptibility to Adversarial Perturbations:}} We test the adversarial robustness of popular pre-trained ImageNet Classifiers against 3 different attacks: FGSM, PGD and AutoAttack. The results for these experiments are provided in  Table~\ref{tab:ImageNetFGSM}. In the table, we use FGSM by varying its perturbation scale  $\epsilon$ from the set ${0, 0.25, 1}$, where $0$ indicates clean images. The image dynamic range is $[0, 255]$. For the reported top-1 accuracy, percentage reductions for CSL methods SimCLR, MoCov3 and SwAV are much larger than for the supervised model. The difference is particularly large for the weaker perturbations, which indicates the higher sensitivity of the model predictions. The results align well with the insights in Section~\ref{sec:explanation}. The observation also holds for the variants of the stronger PGD attack. We provide results for the standard $\ell_{\infty}$ and $\ell_2$ variants of the algorithm, performing 40 iterations, which is commonly used in the literature. Table~\ref{tab:ImageNetFGSM} points to the higher relative adversarial sensitivity of the self-supervision models. DINO and BYOL which do not utilize negative pairs during training, demonstrates higher robustness than the CSL methods SimCLR and MoCov3, which provides further evidence for our hypothesis. We also note that for the pre-trained checkpoints, ViT models are more robust than ResNet. This observation has previously been reported in the literature~\cite{naseer2021intriguing}. However as other works~\cite{pinto2022impartial, bai2021transformers} have suggested that the root cause cannot be attributed to differences in architecture, we do not make any claim about the relative robustness of transformers and CNNs.

 \noindent{\textbf{Susceptibility to Natural Image Corruptions:}} We also employ the ImageNet-C dataset \cite{ImageNetC} to analyze the robustness of models to more primitive transformations, e.g., blurring and noise addition. ImageNet-C includes these perturbations at 5 increasing distortion levels. However, the lowest level is the most relevant to our analysis because we are concerned with the higher local sensitivity of models. We summarize those results in Figure.~\ref{fig:drops}. On average SimCLR \& SwAV experience a 33\% and 12\% higher drop in accuracy relative to the supervised model, while BYOL performs about equally as well as supervised. The results establish higher overall sensitivity of the contrastive self-supervised models for 15 image corruption types. The few cases in which SimCLR shows low sensitivity to corruptions are Brightness and Contrast jittering, which are used as augmentations during its training. SwAV is relatively more robust than SimCLR to non-adversarial transformations, which is a natural consequence of its  ability to `cluster' positive samples for a class.

\subsection{Susceptibility of Video Classification} 

To establish that our observations also hold for different types of models, we perform analysis for action recognition, which is a video classification task. Recently, action recognition techniques have started to exploit  contrastive learning. This opens up the avenue of adversarial robustness analysis for the problem.  We use FGSM, PGD and \texttt{AutoAttack} based attacks here and the results are consistent across attacks. We employ an 18-layer R-(2+1)-D model in our experiments. One variant is trained with supervised cross-entropy loss, and other two are trained using contrastive self-supervised learning methods, TCLR~\cite{dave2021tclr} and CVRL~\cite{qian2020spatiotemporal}. We summarize our results on two datasets (UCF101 and HMDB51) in Table~\ref{tab:UCFFGSM} which shows that the CSL models also gets affected more strongly by the attack as compared to the supervised models. This is true despite self-supervised models considerably outperforming the supervised model on clean inputs. 




\begin{table*}[h]
\centering
{\small
\bgroup
\def\arraystretch{1.0}
\resizebox{\textwidth}{!}{%
\begin{tabular}{l|c|cc|cc}
\hline

\hline

\hline\\[-3mm]
& \multicolumn{1}{r}{\textbf{Attacks $\rightarrow$}}  & \multicolumn{2}{c}{\textbf{FGSM}-$\ell_\infty$} & \multicolumn{2}{c}{\texttt{\textbf{AutoAttack}}-$\ell_\infty$}   \\
\cline{2-6}
\textbf{Pre-Training}               & $\epsilon$ = \textbf{0} & $\epsilon$ = \textbf{8/255}      & $\epsilon$ = \textbf{16/255} & $\epsilon$ = \textbf{1/255}      & $\epsilon$ = \textbf{2/255}   \\
\hline
Supervised Contrastive (SupCon)      & 95.5   & 38.8 \decralgn{59\%} & 31.8 \decralgn{67\%} & 24.3 \decralgn{74\%} & 11.5 \decralgn{88\%}  \\
Self-Supervised Contrastive (SimCLR) & 92.7   & 26.8 \decralgn{71\%} & 13.4 \decralgn{86\%} & 13.7 \decralgn{85\%} & 4.3 \decralgn{95\%}  \\
\hline
SimCLR + FNC~\cite{huynh2022boosting}\large{\textsuperscript{\textdagger}}    & 94.9 & 28.6~\decralgn{70\%}~\increase{8\%} & 16.3~\decralgn{83\%}~\increase{16\%} & 14.7~\decralgn{85\%}~\increase{1\%} & 6.6~\decralgn{93\%}~\increase{27\%} \\
SimCLR + IFND~\cite{chen2021incremental}\large{\textsuperscript{\textdagger}} & 95.1 & 29.2~\decralgn{69\%}~\increase{16\%} & 18.8~\decralgn{80\%}~\increase{32\%} & 15.1~\decralgn{84\%}~\increase{9\%} & 7.1~\decralgn{93\%}~\increase{29\%} \\

\hline
\textbf{Ours (SimCLR + Adaptive FNC)} &&&&& \\
Fixed Threshold ($\rho = 0.9$)  & 93.1   & 30.1 \decralgn{68\%} \increase{25\%} & 21.3 \decralgn{77\%} \increase{53\%} & 17.4 \decralgn{81\%} \increase{25\%} & 7.5 \decralgn{92\%} \increase{43\%} \\
Adaptive Threshold ($\rho = 0.9 \rightarrow 0.5$)    & 93.2   & 31.3~\decralgn{66\%} \increase{42\%} & 21.6~\decralgn{76\%} \increase{53\%} & 19.1~\decralgn{79\%} \increase{55\%} & 8.6~\decralgn{90\%} \increase{70\%}  \\
\rowcolor{LightCyan}
\textbf{Adaptive Threshold ($\rho = 0.99 \rightarrow 0.7$) }  & \textbf{93.5}   & \textbf{33.6}~\decralgn{64\%} \increase{58\%} & \textbf{24.9}~\decralgn{73\%} \increase{68\%} & \textbf{19.3}~\decralgn{79\%} \increase{58\%} & \textbf{8.7}~\decralgn{90\%} \increase{72\%}  \\

\hline
\end{tabular}}
\egroup
}
 \caption{Robustness improvement with our method. Top-1 accuracy for CIFAR-10 under FGSM and \texttt{AutoAttack}. The first two rows provide results without false negative removal. Drop in accuracy under attack is reported in \textcolor{red}{$\downarrow$red}, percentage of gap closed w.r.t.~supervised contrastive learning is indicated in \textcolor{forestgreen}{\textbf{$\uparrow$green}}. \hl{Best Result} closes $>$\textbf{58\%} of the gap. \large{\textsuperscript{\textdagger}}- re-implementation.}
\label{tab:ROBUST}
\end{table*}

\section{Enhancing CSL Robustness} 
\label{sec:robustify}

\begin{table*}[t]
\centering
{\small
\bgroup
\def\arraystretch{1.0}
\resizebox{\textwidth}{!}{%
\begin{tabular}{l|c|cccc|cc}
\hline

\hline

\hline
 & & \multicolumn{2}{c}{\textbf{PGD}-$\ell_\infty$} & \textbf{PGD}-$\ell_2$ & \textbf{PGD}-$\ell_1$ & \multicolumn{2}{c}{\texttt{\textbf{AutoAttack}}-$\ell_\infty$}   \\
\cline{2-8}
\textbf{Pre-Training} & $\epsilon$ = 0 & $\epsilon$ = $\frac{8}{255}$ & $\epsilon$ = $\frac{16}{255}$ & $\epsilon$ = 0.25 & $\epsilon$ = 12 & $\epsilon$ = $\frac{8}{255}$ & $\epsilon$ = $\frac{16}{255}$ \\[1mm] 
\hline
Supervised   & 95.5   & 0.0 & 0.0 & 24.8 & 25.4 & 0.0 & 0.0 \\
Self-Supervised  & 92.7   & 0.0 & 0.0 & 17.1 & 21.1 & 0.0 & 0.0 \\
\hline 
RoCL~\cite{kim2020adversarial}        & 86.0   & 43.6 & 11.4 & 70.9 & 80.0 & 40.8 & 11.2\\
  
\rowcolor{LightCyan} \textbf{Ours}~\textit{(Augmented-RoCL)} & 87.9   & 45.9~\increase{5.3\%}  & 13.2~\increase{15.8\%}  & 72.8~\increase{2.7\%}  &  82.1~\increase{2.6\%} & 43.1~\increase{5.6\%} & 12.1~\increase{8.0\%} \\
\hline
ACL~\cite{Jiang2020}          & 86.2   & 41.2                   & 12.1                   & 72.3                  & 80.7                   & 39.8 & 10.2 \\
\rowcolor{LightCyan} \textbf{Ours}~\textit{(Augmented-ACL)} & 87.9   & 42.5~\increase{3.1\%}  & 13.2~\increase{9.5\%}  & 75.9~\increase{5.0\%} & 83.5~\increase{3.5\%}  & 41.3~\increase{3.7\%} & 10.8~\increase{5.5\%} \\
\hline
\end{tabular}}
\egroup
}
 \caption{Top-1 accuracy of adversarially trained CIFAR-10 models under PGD attack and \texttt{AutoAttack}. Attack strength $\epsilon$ is expressed in terms of $\ell_{\infty}$, $\ell_{2}$ and $\ell_{1}$ norms. The first two rows provide results without adversarial training. Robust models are trained with PGD $\ell_{\infty}$ adversary. Percentage performance gain of our false negative removal augmented methods over adversarially trained RoCL (median gain across attacks \textcolor{forestgreen}{\textbf{5.5\%}}) and ACL (median gain \textcolor{forestgreen}{\textbf{4.4\%}}) is in \textcolor{forestgreen}{$\uparrow$green}~. 
 }
\label{tab:advtrain}
\end{table*}

We have thoroughly established that self-supervised contrastive learning is more sensitive to adversarial inputs than  supervised learning. Our analysis in Section~\ref{sec:explanation} points to the presence of false negative instances in the training data as the major cause of this higher sensitivity. Thus, detecting and removing those can potentially improve the model robustness. In this section we seek to demonstrate that the CSL process itself can be modified to address the issue. Our modifications can further also be applied to  adversarial training methods for CSL and achieve gains in robustness without adding any significant computational overhead to the process.

\subsection{Adaptive False Negative Cancellation} 

Identifying instances that belong to the same semantic class is not straightforward in the absence of label information. By definition, false negatives must share similar features as the true positives. This suggests that we can decide on a suspect false negative by measuring the similarity between a sample's representation to that of the anchor in our mini-batch. Recall from Section~\ref{sec:related} that two prior works have proposed variants of CSL utilizing false negative removal strategies with the goal of improving performance on clean data. False Negative Cancellation (FNC)~\cite{huynh2022boosting} proposes cancelling all negatives which have a cosine similarity higher than a fixed threshold w.r.t. to the anchor and also additionally top-$k$ similar negatives from each batch. Incremental False Negative Detection (IFND)~\cite{chen2021incremental} on the other hand utilizes offline unsupervised clustering to propagate psuedo-labels. The key benefit of FNC is that it does not increase the training time over simple CSL significantly, however unlike IFND, it utilizes a fixed threshold and $k$ value and as a result does not adapt to the shifting features learned by the model as training progresses. 

In order to maintain the low computational overhead of FNC, while also accounting for the improvement in representation quality as training progresses, we propose Adaptive False Negative Cancellation (Adaptive FNC). In Adaptive FNC, we replace the standard CSL objective with a modified objective $\mathcal L_{a}(x)$:


\begin{align}
    \mathcal L_{a}(x) \!=\!\! {\underset{}{\mathbb E~}} \! \Big[-\log \frac{sim(x,x_i^+)}{ sim(x,x_i^+)+\sum_i^{N^-}sim(x,x_i^-)}\Big]
    \label{eq:afnc}
\end{align}

Here, the set of selected positive instances $\{x^+_i\}$ and filtered negative instances $\{x^-_i\}$ is chosen on the basis of similarity in the feature space. A dynamic similarity threshold $\rho$ is used, and instances with similarity higher than $\rho$ are selected into $\{x^+_i\}$ and the rest form $\{x^-_i\}$. Since the quality of features is initially low, we begin with a high value of $\rho$ to avoid spurious detections and slowly decrease it as training progresses and similarity scores become more reliable. Removing false negative pairs from the training objective should lead to tighter class clusters and more adversarially robust models. Adaptive FNC is also efficient as detection using the threshold is $O(N)$ during loss computation and does not increase the computational complexity of the training process. Note that unlike prior work FNC we do not use a top-K heuristic to sample additional false negatives. We also avoid False Negative Attraction as detection of false negative pairs during initial phase of training is not accurate and using them as positive pairs negatively impacts robustness. 

\noindent \textbf{Experiments:} We carry out an ablation study with $\rho_{initial} \in \{0.99, 0.9\}$ and $\rho_{final} \in \{0.5, 0.7\}$ in order to determine the best possible setting for our Adaptive FNC method. The results for the two extreme cases are presented in Table~\ref{tab:ROBUST}. A simple baseline with fixed threshold ($\rho_{initial} = \rho_{final} = 0.9$) is also presented to illustrate the benefits of adaptive FNC. Prior works FNC~\cite{huynh2022boosting} and IFND~\cite{chen2021incremental}, focus on improving performance on clean data outperform our method in clean accuracy. However, Adaptive FNC provides substantial improvement in robust accuracy under FGSM and AutoAttack. The improvement here is without adversarial training and with minimal overhead.

\subsection{Augmenting Adversarial CSL} 
\label{sec:advtrain}

Adversarial learning is a widely adopted paradigm for learning robust models in the supervised learning domain~\cite{akhtar2021advances}. We demonstrate that our Adaptive FNC technique can readily augment adversarial learning in the CSL domain. To that end, we enhance Robust Contrastive Learning (RoCL)~\cite{kim2020adversarial} and Adversarial Contrastive Learning (ACL)~\cite{jiang2020robust} methods with our Adaptive FNC technique. Here, it is also pertinent to mention that referring to \cite{kim2020adversarial}, \cite{carmon2019unlabeled},  \cite{chen2020adversarial}, Hendrycks et al.~\cite{hendrycks2019} alluded to the idea that self-supervision can help in adversarial robustness. Our previous findings provide evidence against this idea for CSL methods. This makes our contribution towards the enhancement of adversarial contrastive learning even more relevant. Discussion on the details of enhancing the RoCL and ACL methods with our technique are provided in the supplementary material.

\noindent \textbf{Experiments:} To evaluate the performance gain achieved by A-RoCL and A-ACL, we followed \cite{kim2020adversarial} and performed adversarial training with PGD $\ell_\infty$-norm bounded adversary. The model robustness is evaluated for $\ell_\infty$, $\ell_1$, $\ell_2$ PGD and $\ell_\infty$ \texttt{AutoAttack}. Results are averaged over 5 training runs and summarized in Table~\ref{tab:advtrain}. Our Augmented-RoCL and Augmented-ACL consistently improve performance gain over RoCL~\cite{kim2020adversarial} and ACL~\cite{jiang2020robust} baselines.

\section*{Conclusion}

In this paper, we investigate the adversarial susceptibility of Contrastive Self Supervised Learning (CSL) trained models. We empirically verify that the use of negative pairs during CSL training is linked to susceptibility. This link is then used to enhance the robustness of CSL and adversarial CSL.

\section*{Acknowledgements}

This work was supported in part by the Defense Advanced Research Projects Agency (DARPA) under Agreement HR00112090095. Dr. Naveed Akhtar is the recipient of Office of National Intelligence, National Intelligence Postdoctoral Grant (project number NIPG-2021-001) funded by the Australian Government. Professor Ajmal Mian is the recipient of an Australian Research Council Future Fellowship Award (project number FT210100268) funded by the Australian Government.

The authors would also like to thank Ishan Dave for fruitful discussions on video contrastive learning methods.

\bibliography{aaai23}

\end{document}